\documentclass[conference]{IEEEtran}
\IEEEoverridecommandlockouts
\usepackage{cite}
\usepackage{amsmath,amssymb,amsfonts}
\usepackage{algorithmic}
\usepackage{graphicx}
\usepackage{textcomp}
\usepackage{xcolor}
\usepackage{ulem}
\usepackage{comment}
\usepackage{caption}
\usepackage{subcaption}
\usepackage{multirow}
\usepackage{multirow}
\usepackage{tabularx}
\usepackage{array}
\usepackage{float}
\def\BibTeX{{\rm B\kern-.05em{\sc i\kern-.025em b}\kern-.08em
    T\kern-.1667em\lower.7ex\hbox{E}\kern-.125emX}}

\colorlet{FigColor}{green!30!blue!70!black!100!}
\colorlet{darkBlue}{blue!30!black!70!}

\begin{document}

\title{Obscured Wildfire Flame Detection By Temporal Analysis of Smoke Patterns  Captured by Unmanned Aerial Systems}

\author{\IEEEauthorblockN{Uma Meleti}
 \IEEEauthorblockA{\textit{School of Computing} \\
\textit{Clemson University}, Clemson, SC, USA \\
umeleti@clemson.edu}
\and
\IEEEauthorblockN{Abolfazl Razi}
\IEEEauthorblockA{\textit{School of Computing} \\
\textit{Clemson University},Clemson, SC, USA \\
arazi@clemson.edu}}

\maketitle

\begin{abstract}

This research paper addresses the challenge of detecting obscured wildfires (when the fire flames are covered by trees, smoke, clouds, and other natural barriers) in real-time using drones equipped only with RGB cameras. We propose a novel methodology that employs semantic segmentation based on the temporal analysis of smoke patterns in video sequences. Our approach utilizes an encoder-decoder architecture based on deep convolutional neural network architecture with a pre-trained CNN encoder and 3D convolutions for decoding while using sequential stacking of features to exploit temporal variations. The predicted fire locations can assist drones in effectively combating forest fires and pinpoint fire retardant chemical drop on exact flame locations.
We applied our method to a curated dataset derived from the FLAME2 dataset that includes RGB video along with IR video to determine the ground truth. 
Our proposed method has a unique property of detecting obscured fire and achieves a Dice score of 85.88\%, while achieving a high precision of 92.47\% and classification accuracy of 90.67\% on test data showing promising results when inspected visually. 
Indeed, our method outperforms other methods by a significant margin in terms of video-level fire classification as we obtained about 100\% accuracy using MobileNet+CBAM as the encoder backbone. 
\end{abstract}


\section{Introduction}
Wildfires have become prevalent and destructive in many parts of the world. Regardless of the cause, wildfires can have severe consequences, including the loss of human lives, destruction of property, disruption of wildlife, food production, and crop supply chain as well as significant environmental damage.  Once a wildfire starts, there are various ways to monitor and control these wildfires, including observation towers, direct human intervention, satellite imaging, and manned aircraft. Using drones is one of the most efficient ways of fire monitoring for its low operation cost, customizable sensing and imaging, flexible operation, and ease of use in harsh environments by advanced flight control features and partial autonomy (e.g., safe auto-landing). 

One of the most significant advantages of using drones in wildfire fighting is the ability to gather real-time data about the fire’s behavior. Equipped with cameras and other sensors, drones can fly over the fire and capture relevant information. Drones can also be equipped with water tanks or fire extinguishers containing fire-retardant chemicals, such as carbon dioxide (CO2), potassium bicarbonate (KHCO3), or evaporating like bromochlorodifluoromethane (CF2ClBr) to be dropped on hot spots to create fire breaks. The efficient utilization of drones will significantly advance controlling these fires.

Targeting places where actual burning is happening, instead of blindly spraying fire retardant gases everywhere, will help set these fires off quickly and efficiently. Finding out these burning places in real-time is very difficult, especially when fire flames are obscured by thick smoke. An infrared camera can help find these hidden fires, but IR cameras are expensive.

We devised a methodology that uses an RGB camera feed, analyzes the video frames sequentially, and detects the obscured fires using temporal features, such as smoke patterns, extracted from video frames. We show that such features can be indicative of 
the fire's exact location and temporal behavior. We have structured the problem as semantic segmentation of the obscured 
fire by analyzing the sequence of frames in the video. A deep convolutional neural network (CNN) is designed that uses a pre-trained CNN architecture as an encoder to extract features of video frames and passes sequentially stacked features to the decoding stage that uses 3D convolutions \cite{tran2015learning}. to analyze these features and predict burning location. UAVs and Drones can use this predicted information for more guided and informed fire monitoring and control. 

Specifically, we propose a novel method for obscured fire detection that can be used for other applications beyond forest fire management. 
To this end, we curated a dataset from the existing FLAME2 \cite{shamsoshoara2021aerial} dataset 
by selecting part of video frames from the original data where the drone is stationary and there is high synchronization with the IR images to avoid misalignment errors. We also use the corresponding IR videos to extract ground truth for our task by performing a series of Image Processing operations. We have visually verified the correspondence of RGB video with the processed IR video for synchronization. 
We highlight the unique features of our method compared to previous methods, then proceed by elaborating on the details of the generated dataset and its preparation method. We then elucidate the details of the proposed deep learning (DL) architecture and analyze the obtained results.

\section{Related work}

The previous works on fire detection are mainly based on image-based techniques such as classification, object detection, and semantic segmentation of visible fire imagery. Wonjae Lee et al. proposed a wildfire detection system that classifies the presence of fire in images and evaluated the performance of AlexNet, GoogLeNet, and VGG, along with their modified variants \cite{7889305}. Zhentian et al. have trained YOLOV3 for fire detection and reported a recognition rate of 91\% \cite{9163816}. An ensemble-based object detection method using YOLOV5 EfficientDet is proposed in \cite{xu2021forest} for detection and EfficientNet to capture global information about the fire. Their study showed a decrease in false positive rate by 51.3\% on three public datasets. Yo Zhao et al. have proposed a deep learning architecture, called Fire-Net by stacking convolutional and pooling layers for fast localization and segmentation of fire in aerial images with an accuracy of 98 \% on standard 'UAV\textunderscore Fire' dataset \cite{zhao2018saliency}. A similar method is proposed in \cite{zhang2022forest} and \cite{wang2022real} by using a deep learning architecture to classify frames into "fire with smoke", "fire with no smoke", and "no fire" using FLAME datasets \cite{shamsoshoara2021aerial,swyw-6j78-22}. However, most of these methods are limited to images, where real-time data for fire detection will be mostly in terms of video feed. Further, video feeds contain temporal patterns of smoke that facilitate locating the origin of smoke which is the fire location, and distinguishing them from clouds and other white patterns. This concept is used as a key idea in our method to detect obscured fire positions.

A few works take a slightly different approach and analyze fire images path by patch instead of one-shot analysis of the entire image. For instance, a CNN-based deep learning method is proposed in \cite{zhang2016deep} which classifies the image first and then performs a patch-level analysis to offer more detailed information. They applied their method to video frames to perform patch-wise detection and reported a 97 \% detection accuracy on their own dataset. Still, this method does not consider the temporal relationship for fire classification since it treats video frames as still images. In a similar work, Gwangsu Kim et al. proposed an algorithm that collects features of video frames using a pre-trained VGG and stacks them together to be passed through a series of fully connected layers to classify the presence of fire in the video clips. However, this method restricts fire classification only during a visible fire deemed inefficient in capturing obscured fire.

Anshuman et al. \cite{dewangan2022figlib} have proposed SmokeyNet - a deep learning algorithm that offers stacking CNN, LSTM, and Vision Transformer to detect smoke in the video feed captured by stand-alone cameras. However, this method is not directly applicable to aerial imagery. 
Some other research works take advantage of Infrared (IR) cameras for more accurate fire positioning. 
For instance, Chi Yuan et al. proposed an algorithm that uses brightness and motion clues with histogram-based segmentation and optical flow to segment fire in IR images \cite{yuan2017fire}. Another example is Norsuzial et al.'s work that offers an Image processing-based approach to convert IR images to YCbCr color and use a wavelet analyzer to detect fire \cite{ya2021image}. Also, a DL architecture is proposed in \cite{9953997} that analyzes dual-feed imagery captured by side-by-side RGB and IR cameras for precise fire positioning. 
Although these methods yield high accuracy taking advantage of thermal information captured by IR cameras, they incur an extra monitoring cost for their reliance on pricy IR cameras. Also, they are not suitable for processing existing drone-based and satellite-based datasets that include only RGB imagery. 
In contrast to all the above, our method uses only RGB videos for detecting both visible and obscured fire flames in an economical way.

\section{Data Preparation}

In this study, we have used the publicly available FLAME2 dataset \cite{swyw-6j78-22}, which consists of 7 video pairs of RGB and corresponding infrared heat maps. Out of those, we have employed five relevant videos in our simulations because these videos consist of both visible fire and obscure fire, appropriate for our test. The videos were taken in a planned burning region, consisting of information on forest burning with smoke. The drone move around the place, covering different parts of the woods. For experimentation purposes, we have carefully cropped the parts of the video where the camera is relatively stationary and there is a high alignment between the RGB and IR camera viewpoints. The selected video segments are split into 
clips of 20 consecutive frames to train our deep neural network, where each clip is considered a training sample. 

\section{Method}

\subsection{Data Pre-Processing: Using IR to Label RGB images}

The IR images consist of heat maps corresponding to the temperature of different regions on the image. The place where the fire is present generally has a high temperature, and the pixel values in the heat map are closer to the maximum. We have extracted the ground truth for training data where the fire is present by processing the IR image with a series of hand-crafted image processing methods. The set of operations performed on the IR image is shown below.

\vspace{10pt}
IR Image $\rightarrow$ Smooth Image $(5 \times 5)$ $\rightarrow$ Hard Thresholding $\rightarrow$ Dilation $(5 \times 5$, 2 times$)$ $\rightarrow$ Fill (flood fill) $\rightarrow$ Erosion $(5 \times 5$, 1 time$)$ $\rightarrow$ Remove small objects $(200$ px$)$ $\rightarrow$ Ground Truth.

We initially smooth out the image using a low-pass filter to remove noise, then use hard thresholding to select the regions of high-temperature values corresponding to fire. The resulting image is dilated to fill the small spaces and to make the fire boundary smooth; the spaces not filled in the previous operations are filled using flood fill that will result in a complete blob of fire location. The image is eroded to reverse the effect of dilation applied earlier. Small blobs that are likely to be representative of noise are removed.

Note that we use IR images to identify ground truth fire locations and train the model, but in runtime (new monitoring tasks), we only use RGB images, so our method does not require expensive IR cameras on site.

\begin{figure}[!ht]
\begin{center}
\setlength\belowcaptionskip{-2\baselineskip}
\centerline{\includegraphics[width=1\columnwidth]{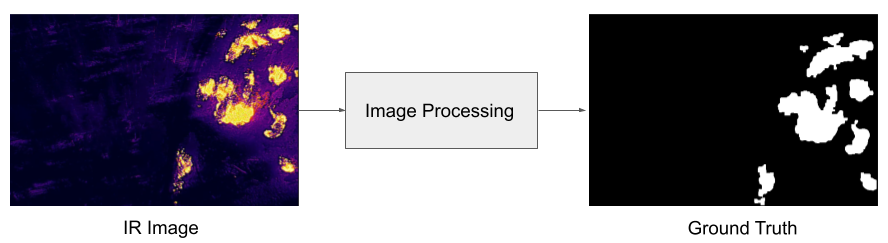}}
\caption{Obtaining Ground truth fire locations from IR images.}
\label{fig:atp}
\end{center}
\end{figure}

\subsection{Ground Truth Approximation}

The main goal is to take a sequence of input frames and predict where the fire hides. Since we obtain ground truth from the IR camera feed, every video frame has a pixel-wise map representing the ground truth. However, we need to define a single ground truth for each sample (i.e. 20-frame clip). To this end, we have approximated the ground truth by applying majority voting to the pixel labels obtained from the 20 IR video frames.
More specifically, we have
\begin{equation}
    \text{Final Label}(L_{i,j}^*) = \text{majority\_class}(L_{i,j}^1, L_{i,j}^2, \ldots, L_{i,j}^\text{seq\_len}), 
\end{equation}
where $(i,j)$ determines the pixel location, and the postscript is the frame number within the clip. In our case, the label is binary, so the class is either 0 or 1.

\begin{figure}[!ht]
\begin{center}
\setlength\belowcaptionskip{-2\baselineskip}
\centerline{\includegraphics[width=1\columnwidth]{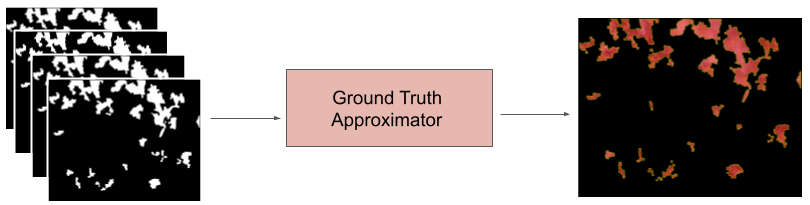}}
\caption{Ground Truth Approximation}
\label{fig:atp}
\end{center}
\end{figure}

\subsection{Network Architecture}
We have presented the overview of our architecture in Fig 3. The architecture consists of a pre-trained encoder that encodes features from the video frames along with a 3D decoder that decodes information from the volume of features.

We have used VGG16 \cite{simonyan2015deep} as the encoder and pass a sequence of video frames, then collect each frame's features at different resolutions of the encoder and stack them to pass it to the 3D Decoder, which processes these volumes of features to predict the segmentation map of the hidden fire.

The Decoder consists of two parts; the first part decodes information in both the image and time axis, but more emphasis is put on summarizing the semantic information of the image that will be used by the second part. This enforces the decoder to focus more on capturing the relationship between the semantic features between the frames in the time axis.

\begin{figure*}[htbp]
    \centering
    \includegraphics[width=\textwidth, height=0.5\textwidth]{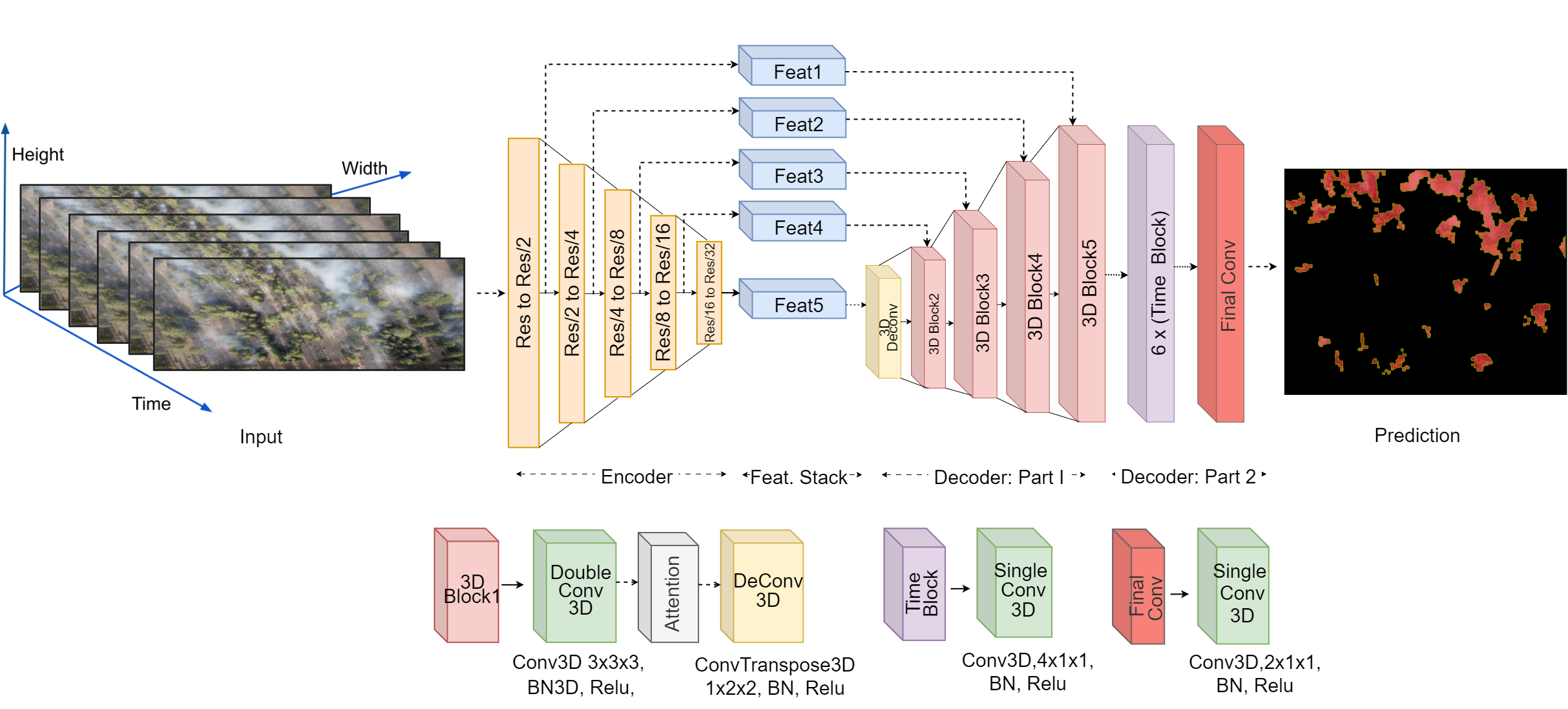}
    \caption{The overall architecture of the proposed deep learning network for obscured fire detection.} 
    \label{fig:example}
\end{figure*}

\subsection*{Decoder: Part-1}

The design of the first part of the Decoder is inspired by the U-Net architecture\cite{ronneberger2015unet}, where features from multiple resolutions are merged with the Decoder. We extracted features of each frame at resolutions of (HxW)/2, (HxW)/4, (HxW)/8, (HxW)/16, and (HxW)/32, where H, W are the height and width of the input frame. The extracted features are stacked for a sequence of frames. At each resolution, the volume of components is processed by a 3x3x3 convolution block followed by an attention block, which learns the most informative feature representations while reducing the feature space. This architecture retains the dimension of the input 
in both the feature domain and time axis. 

A deconvolution layer is applied to the bottleneck of the encoder with a (1x2x2) Transposed convolution, which upsamples the feature map and increases the resolution by a factor of 2. The upsampled feature map is then concatenated and fed into the convolution, attention, and deconvolution layers. This process is repeated until the output resolution becomes exactly equal to the input resolution. The ultimate output dimension of this block is (batch\textunderscore size, n\textunderscore classes, seq\textunderscore length, H x W).

\subsection*{Decoder: Part 2}

The Decoder2 consists of a series of Time blocks and a final convolution layer. The Time block consists of consecutive operations of 3D Convolution, Batch Normalization\cite{ioffe2015batch}, and ReLU Activation. We have chosen a kernel size of (4x1x1) for the convolution so that every block captures information of 4 consecutive frames. 
Here we have used a 1x1 kernel size in the feature space and a size of 4 in time-space; the idea is to give more emphasis in time-space than feature space. 
In our experiments, we have considered a sequence length of 20 frames as input to predict the output; we added 6 Time blocks that will reduce the feature space in the time dimension and reach a resolution of (2 x classes x H x W) and a final convolution layer is added to reduce to the final resolution of 1 x classes x H x W.

\subsection{Loss Function}

We have used Dice loss to measure the alignment between the ground truth and the detected fire regions by the architecture to train the network. The Dice coefficient measures the alignment (similarity) 
between two corresponding segments by computing the overlap coefficient. The overlap coefficient ranges from 0 to 1, with 1 indicating a perfect match between the two segments. The Dice loss function is defined as one minus the Dice coefficient, with the objective of minimizing the loss during training. More specifically, we have

\begin{align}
\label{eq:dice}
\nonumber 
DiceLoss &= 1- \frac{2 \times \text{Intersection}}{\text{Predicted} + \text{Ground Truth}} \\
&=1 - \frac{2 \sum_{i=1}^{n}p_i g_i + \epsilon}{\sum_{i=1}^{n}p_i^2 + \sum_{i=1}^{n}g_i^2 + \epsilon},
\end{align}
where $p_i$ and $g_i$ are the predicted and ground truth segmentation masks, respectively, for the $i$-th pixel in the image. The summations are taken over all $n$ pixels in the image. The $\epsilon$ term is a small constant added to the denominator to prevent division by zero.

\section{Experiments}

In this section, we present the simulation results using the Flame2 dataset, which consists of RGB and IR images; we have trained with pre-processed videos as explained in the Data Preparation step. We have used 354 videos for training and 155 videos for testing. All the training and testing videos are independent and non-overlapping frames.

\subsection{Training}
Our model is implemented in PyTorch \cite{ronneberger2015unet} using a Linux machine with Tesla A-100 40 GB GPU.
The models were tuned for the best hyperparameters. We used a step-learning rate with an initial learning rate of 1e-2 with Adam Optimizer \cite{kingma2014adam}. The models were trained for 300 epochs with a batch size of 5.

\subsection{Inference}

The inference in real-time, where videos are lengthy, is made by taking frames at a window size of 20 and sliding the window over the video at a stride of one.

\begin{figure}[!ht]
\begin{center}
\setlength\belowcaptionskip{-2\baselineskip}
\centerline{\includegraphics[width=0.95\columnwidth]{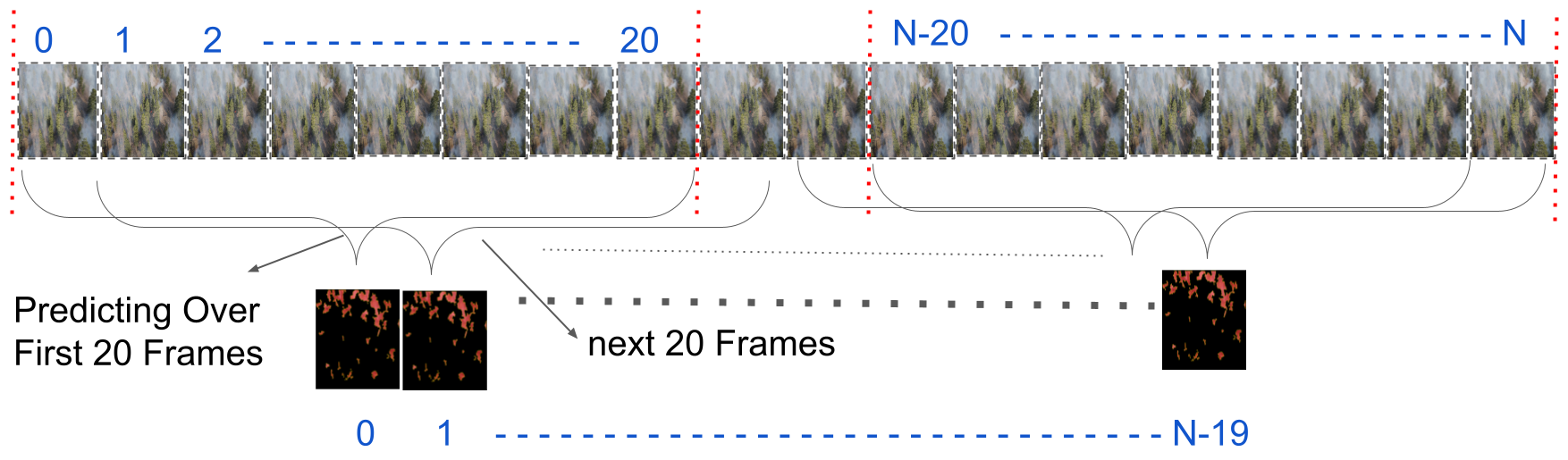}}
\caption{Inference method: each output map is the result of sequential processing of 20 preceding input frames.}
\label{fig:atp}
\end{center}
\end{figure} 

\begin{figure*}[htbp]
    \centering
    \includegraphics[width=\textwidth]{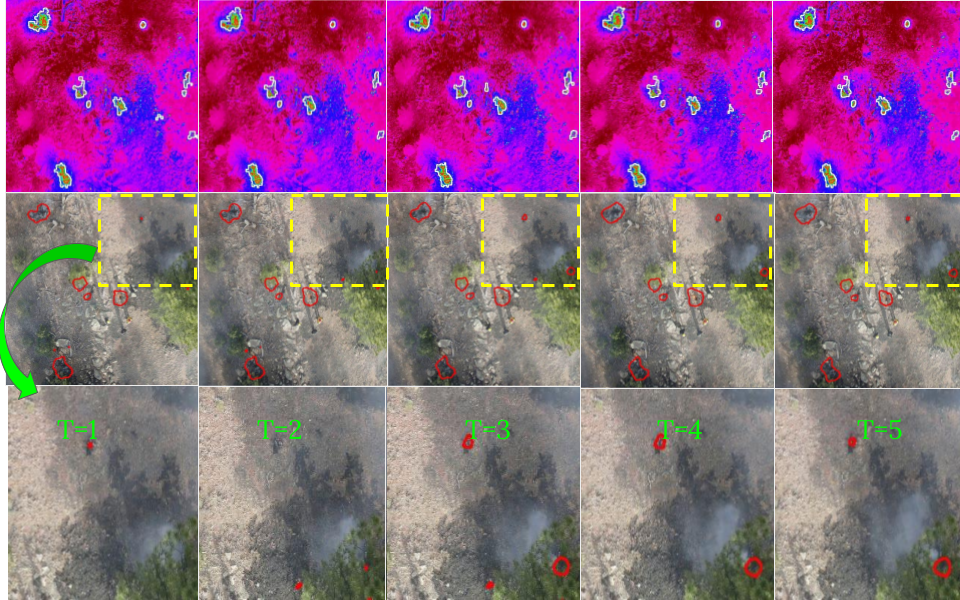}
    \caption{Sample output of Model applied on consecutive frames from left to right; Top - IR with ground truth (white boundary), Middle - Prediction on RGB Image, Bottom - Zoomed RGB. (Red - Prediction boundary).}
    \label{fig:qual_res}
\end{figure*}


\begin{figure}[!ht]
\begin{center}
\setlength\belowcaptionskip{-2\baselineskip}
\centerline{\includegraphics[width=1\columnwidth]{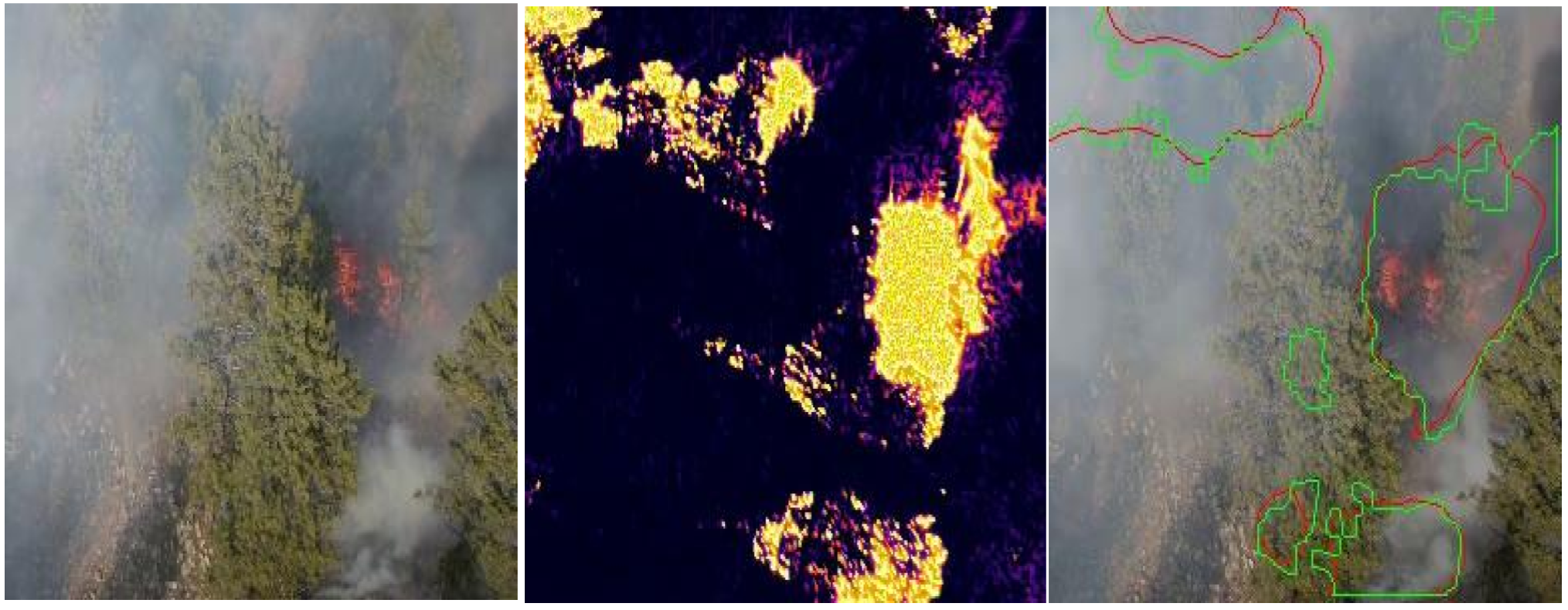}}
\caption{Left RGB, Middle IR, Right Ground Truth (Green), Prediction (Red) annotated}
\label{fig:overlay}
\end{center}
\end{figure} 

\subsection{Evaluation Metrics}


We are using a set of metrics to evaluate the quality of fire detection. We used the Dice score (presented in (\ref{eq:dice})) to assess the alignment quality between ground truth and the detected fire region. 
This assessment is particularly important when part of the fire is obscured by thick smoke to demonstrate to what extent our model is capable of detecting such fire regions. 


Another metric we used is blob-wise precision to ensure the correctness of our predictions; it is calculated by taking each blob in the ground truth and prediction, and if the intersection area is greater than 30\%, considering it as a True positive, else a false positive. Precision is calculated by using the formula.

\[
\text{Precision} = \frac{\text{True Positives}}{\text{True Positives} + \text{False Negatives}}
\]

And also, calculated clip-level classification accuracy to evaluate the video with fire is being classified as fire or not fire. This is calculated by counting the number of fire spots in the video and comparing it with the prediction; if more than 30\% spots are predicted we will classify the video as fire, else non-fire.

\subsection{Quantitative Results}

The Quantitative results are shown in Table \ref{tab:model_performance}. We examined four different types of backbones (VGG16\cite{simonyan2015deep}, ResNet\cite{he2015deep}, EfficinetNet\cite{tan2019efficientnet}, MobileNet\cite{howard2017mobilenets}) and two different types of attention modules. One is Spatial and Channel Squeeze \& Excitation Blocks (ScSE) \cite{roy2018recalibrating} and another is Convolutional Block Attention Module (CBAM) \cite{woo2018cbam}. On overall backbones, VGG16 has shown the highest performance in terms of fire region detection alignment (dice), but Efficinetb0 with ScSE has shown a similar dice score and also has superior performance in blob-wise precision. ResNet18+CBAM and MobileNet+CBAM has shown 100\% percent classification accuracy. This shows that our architecture is flexible and different types of pre-trained architectures can be employed as the encoder backbone. 

\begin{table*}[t]
  \caption{Model Performances with various backbones and attention types}
  \centering
  \begin{tabular}{|l|c|c|c|c|c|c|c|}
    \hline
    \multirow{2}{*}{Backbone+Attn.Type} & \multirow{2}{*}{Parameters} & \multicolumn{2}{c|}{Mean Dice} & \multicolumn{2}{c|}{Precision} & \multicolumn{2}{c|}{classification Accuracy} \\
    \cline{3-8}
     & & Train & Test & Train & Test & Train & Test \\
    \hline
    VGG16+CBAM & 26M & 85.44 & 84 & 77.84 & 71.32 & 96.32 & 98.71 \\
    VGG16+ScSE & 26M & \textbf{90.21} & \textbf{86.32} & 93.36 & 72.73 & 94.91 & 99.35 \\
    ResNet18+CBAM & 18M & 82.5 & 82.24 & 88.68 & 85.61 & 99.71 & \textbf{100.00} \\
    ResNet18+ScSE & 18M & 77.82 & 77.90 & 76.83 & 76.63 & 79.96 & 80.64 \\
    Effb0+ScSE & 10M & 86.01 & 85.88 & \textbf{94.78} & \textbf{92.42} & 90.67 & 90.67 \\
    Effb0+CBAM & 10M & 81.00 & 81.04 & 81.46 & 82.46 & 95.19 & 94.83 \\
    Effb1+ScSE & 12M & 73.09 & 73.21 & 74.26 & 74.38 & 95.76 & 95.48 \\
    Effb1+CBAM & 12M & 80.36 & 80.19 & 85.66 & 83.57 & 97.74 & 96.13 \\
    MobileNet+ScSE & 12M & 72.88 & 73.47 & 61.54 & 60.02 & \textbf{100.00} & 99.35 \\
    MobileNet+CBAM & 12M & 80.55 & 80.23 & 78.75 & 76.88 & \textbf{100.00} & \textbf{100.00} \\
    \hline
  \end{tabular}
  \vspace{-.2 in}
  \label{tab:model_performance}
\end{table*}

\subsection{Qualitative Results}

Fig. \ref{fig:qual_res} is a sample output of our model applied on consecutive video frames (left to right); the top row presents the IR images from where our ground truth is extracted (annotated in white), and the middle row corresponds to prediction (annotated in red) and last row is zoomed at the annotated region (yellow). at T=1, fire is slowly starting under the tree and the volume of smoke grows gradually in the next frames. Initially, the model does not detect the obscured fire. However, as time passes, the temporal analysis of the growing fire flames enables the model to detect the obscured flame (shown by red colors).

Fig. \ref{fig:overlay} is the output snapped at a particular frame, the left image corresponds to RGB Input and the middle is IR, and the right includes the ground truth fire region (geen line) and the predicted fire region (red line). This image demonstrates that our model detects both visible fire and obscured fire with near-accurate boundaries.

\section{Discussion}
\vspace{-0.1 in}
The quantitative and qualitative performance of our model yields promising results using various backbones exhibiting that the proposed architecture is flexible in adopting various existing and future pre-trained backbones.

We used IR images offered in the Flame2 dataset to determine the ground truth fire regions. It is noteworthy that the temperature values of IR images are calibrated within the frame values and do not reflect the exact temperature, which should be taken into account in the labeling process.

This study can trigger multiple future works. For instance, further research can focus on refining and expanding our methodology, 
considering other environmental factors that may affect fire behavior. The application of our approach in practical scenarios, developing onboard processing software, and integration with existing wildfire management systems can provide valuable insights for future developments.

\section{Conclusion}
In this paper, a novel approach is proposed for detecting obscured fires in real-time using video feeds captured by drones 
equipped only with RGB cameras. The key idea was training a model that treats a video clip as a single sample and processes its video frames sequentially to identify temporal smoke patterns that can be indicative of obscured fires.  To this end, we introduced a new deep-learning architecture that leverages pre-trained CNN architectures and 3D convolutions to create a temporal feature map and use attention modules to predict fire regions by the sequential analysis of video frames. 
We evaluated our method on a curated FLAME2 dataset where the IR videos are used to discover the ground truth fire regions and showed that our method not only improves the fire detection accuracy compared to the state-of-the-art (by achieving near 100\% accuracy), but also demonstrates great success in detecting invisible and covered fire region borders (about 85\% in Dice score) even when they are obscured by trees and smoke patterns. 
This methodology allows utilizing firefighter drones to combat wildfires more efficiently by targeting visible and invisible fire hotspots. 
Also, our method helps detect fire regions precisely without the need for IR cameras (in the test phase) which significantly reduces fire monitoring costs.  

\vspace{-.2 in}
\bibliographystyle{ieeetr}
\bibliography{main} 

\end{document}